\pdfoutput=1

\documentclass[11pt]{article}
\usepackage{amsmath,amsfonts}
\usepackage{ulem}
\usepackage{multirow}

\usepackage[preprint]{acl}

\usepackage{times}
\usepackage{latexsym}

\usepackage[T1]{fontenc}

\usepackage[utf8]{inputenc}

\usepackage{microtype}

\usepackage{inconsolata}

\usepackage{graphicx}

\title{PFME: A Modular Approach for Fine-grained Hallucination Detection and Editing of Large Language Models}

\author{First Author \\
  Affiliation / Address line 1 \\
  Affiliation / Address line 2 \\
  Affiliation / Address line 3 \\
  \texttt{email@domain} \\\And
  Second Author \\
  Affiliation / Address line 1 \\
  Affiliation / Address line 2 \\
  Affiliation / Address line 3 \\
  \texttt{email@domain} \\}

\author{
 \textbf{Kunquan Deng\textsuperscript{1}},
 \textbf{Zeyu Huang\textsuperscript{2}},
 \textbf{Chen Li\textsuperscript{1}},
 \textbf{Chenghua Lin\textsuperscript{3}},
 \textbf{Min Gao\textsuperscript{4}},
 \textbf{Wenge Rong\textsuperscript{1}}\thanks{~~Corresponding Author.},
\\
 \textsuperscript{1}SCSE, Beihang University,
 \textsuperscript{2}ILCC, The University of Edinburgh,
 \textsuperscript{3}DCS, University of Manchester,
\\
 \textsuperscript{4}School of Big Data and Software Engineering, Chongqing University
\\
\texttt{dengkunquan@buaa.edu.cn \quad zeyu.huang@ed.ac.uk}
}

\begin{document}
\maketitle
\begin{abstract}
Large Language Models (LLMs) excel in fluency but risk producing inaccurate content, called "hallucinations." This paper outlines a standardized process for categorizing fine-grained hallucination types~\cite{mishra2024fine-fava} and proposes an innovative framework—the Progressive Fine-grained Model Editor (PFME)—specifically designed to detect and correct fine-grained hallucinations in LLMs.
PFME consists of two collaborative modules: the Real-time Fact Retrieval Module and the Fine-grained Hallucination Detection and Editing Module. The former identifies key entities in the document and retrieves the latest factual evidence from credible sources. The latter further segments the document into sentence-level text and, based on relevant evidence and previously edited context, identifies, locates, and edits each sentence's hallucination type.
Experimental results on FavaBench and FActScore demonstrate that PFME outperforms existing methods in fine-grained hallucination detection tasks. Particularly, when using the Llama3-8B-Instruct model, PFME's performance in fine-grained hallucination detection with external knowledge assistance improves by 8.7 percentage points (pp) compared to ChatGPT. In editing tasks, PFME further enhances the FActScore of FActScore-Alpaca13B and FActScore-ChatGPT datasets, increasing by 16.2pp and 4.6pp, respectively.
\end{abstract}

\section{Introduction}

LLMs have demonstrated an unprecedented level of fluency and natural dialogue capabilities. Despite the significant advancements, these models occasionally generate specious outputs, such as content that does not match the user's input\cite{adlakha2024evaluating}, deviating from the previously generated context~\cite{liu2022token}, or directly contradicting factual knowledge~\cite{min2023factscore}.

In traditional hallucination detection tasks, which are often domain-specific~\cite{devaraj2022evaluating, pagnoni2021understanding, dziri2022origin}, it is usually assumed that a reference data source exists, and any deviations from the original text will be considered hallucinations~\cite{pagnoni2021understanding}. For example, in summarization tasks, any inconsistency between the summary and the document information, or the generation of unverifiable extra information such as phone numbers or addresses, is considered hallucinated.
However, for the LLMs chatbot, the data source can be considered as all the world's knowledge~\cite{li2021addressing}. At the same time, the tolerance of the chatbot is relatively high, and it focuses more on user engagement rather than faithfulness~\cite{ji2023survey}. User engagement mainly depends on factualness. When the additional information generated by the LLMs is irrelevant to the user's input but is factual, this information may still be useful. Therefore, for LLMs, the types of hallucinations that require special attention are those based on factual knowledge of the world instead of detecting faithfulness errors~\cite{zhang2023siren, huang2023survey}.

The handling of hallucinations in LLMs mainly involves two tasks: detection and editing. Existing hallucination detection methods tend to make binary judgments on factual errors in sentences output by large models, simply classifying them as "factual" or "non-factual"~\cite{min2023factscore, gao2023rarr}. To overcome the limitations of traditional binary classification of hallucinations, Mishra et al.~\cite{mishra2024fine-fava} proposed a fine-grained hallucination classification, but there is a lack of standardized judgment processes.

Due to the high cost of training and fine-tuning LLMs, model editing methods have been developed to edit and correct LLM outputs through low-cost methods. Based on the position of editing, existing model editing methods could be divided into three types:
(1) In-context learning (ICL) involves modifying prompts to produce the correct text.~\cite{zheng2023can-ike, brown2020language}
(2) Model Structure-Based Editing introduces additional parameters~\cite{mitchell2022memory-serac, huang2022transformer, li2024pmet} or modifying LLM parameters~\cite{mitchell2021fast-mend, de2021editing-KnowledgeEditor-ke, ha2016hypernetworks}. This method essentially relies on fine-tuning, so knowledge updates are still not flexible enough.
(3) Output-Based Knowledge Editing focuses on editing the output of LLMs~\cite{zhong2023mquake, wang2024deepedit}, but they struggle to overcome persistent knowledge issues~\cite{bi2024decoding}. 
Traditional methods emphasize controlling the model's generation process, but they lack mechanisms for checking the generated results, which may still contain hallucinations, making them unsuitable for hallucination detection scenarios.

In our study, we designed a standardized fine-grained hallucination classification judgment process and developed a Progressive Fine-grained Real-time Fact Editing framework (PFME). This framework consists of two main components: (1) a real-time fact retrieval module that identifies key entities within the document and retrieves the latest factual evidence from trusted data sources based on these key entities; (2) a progressive fine-grained hallucination detection and editing module that splits the document into sentences and progressively identify, locate, and edit fine-grained hallucinations based on the relevant factual evidence.

Following the experiment setting of~\cite{mishra2024fine-fava}, we experimentally demonstrate that PFME, based on Llama3-8B-Instruct~\cite{llama3modelcard}, significantly outperforms the powerful ChatGPT~\cite{openai2024chatgpt} under the same evidence settings in fine-grained hallucination detection tasks: PFME improved the optimal performance of ChatGPT with external knowledge by 8.7 percentage points(pp) in fine-grained hallucination detection tasks. In editing tasks, PFME significantly improves the FActScore of the Alpaca 13B and ChatGPT in generating biographical data sets by 16.2 pp and 4.6 pp, respectively.

To sum up, our contribution is twofold: (1) a standardized definition of fine-grained hallucinations, and (2) a novel framework (PFME) for the fine-grained hallucination detection and editing task.

\section{Related Work}

\paragraph{Classification of Hallucinations}

Hallucination classification in LLMs lacks a unified standard, with different studies proposing varied categorizations. Zhang classify hallucinations into input, context, and fact conflict types~\cite{zhang2023siren}, while Huang distinguish between factual and fidelity hallucinations~\cite{huang2023survey}. Despite this diversity, there's a consensus that deviations from the original text meaning or objective facts constitute hallucinations. In open chatbot systems, where models rely on world knowledge~\cite{li2021addressing, devaraj2022evaluating}, tolerance for fidelity errors is higher, but factual errors are strongly discouraged due to their impact on user engagement and information accuracy~\cite{pagnoni2021understanding}. Hence, LLMs must prioritize detecting factual hallucinations to maintain chatbot effectiveness.

Thus, in studying hallucinations in LLMs, the focus should primarily be on factual hallucinations. However, traditional research has mainly focused on the fidelity of generated text to the source text~\cite{ji2023survey}. Addressing this, Mishra~\cite{mishra2024fine-fava} proposed a classification of fine-grained hallucinations, further subdividing hallucinations into six fine-grained types. We expanded upon this fine-grained hallucination classification work and established standardized judgment criteria for computerized assessment.

\paragraph{Model Editing Methods}

In model editing, two main strategies stand out. The first modifies model parameters directly, facing difficulties in sequential tasks~\cite{meng2022locating-rome,mitchell2021fast-mend, meng2022mass-memit, de2021editing-KnowledgeEditor-ke, ha2016hypernetworks}. The second keeps parameters intact, storing edits in memory for later retrieval~\cite{mitchell2022memory-serac, huang2022transformer, wang2024deepedit, zheng2023can-ike}. Depending on the objective, these edits may involve input, structural components, or output adjustments. While existing methods often excel in specific tasks, they struggle with real-world scenarios. Therefore,~\cite{mishra2024fine-fava}'s fine-grained hallucination detection and editing task presents greater challenges. This task requires editing models to locate and identify hallucinations in the output of LLMs within complex contextual environments and to edit them effectively. Our refinement in sentence-level processing enriches contextual learning, thereby aiding in this task. 

\begin{figure*}[!t]
\centering
\includegraphics[width=0.75\linewidth]{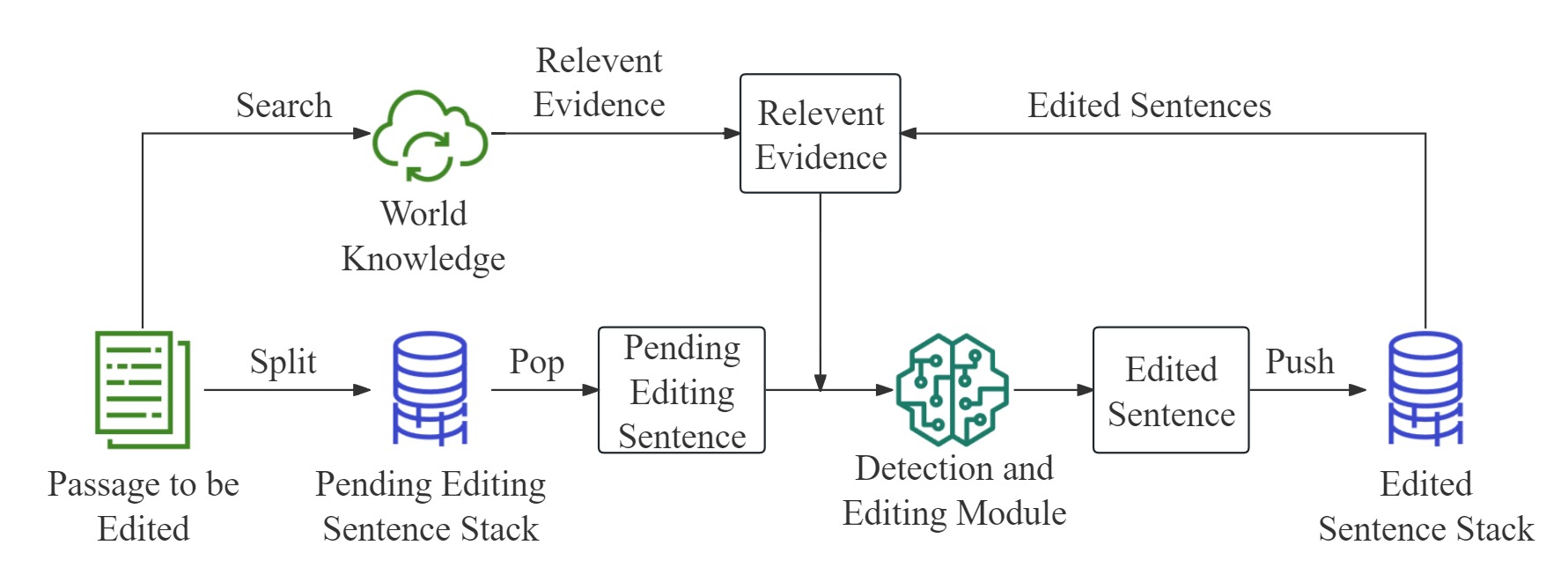}%
\hfil
\caption{Progressive Fine-grained Model Editor Architecture}
\label{fig_architecture}
\end{figure*}

\section{Definition and Judgment Criteria of Fine-grained Hallucinations}

Mishra~\cite{mishra2024fine-fava} classifies fine-grained hallucinations into 6 types: \textit{Entity}, \textit{Relation}, \textit{Contradictory}, \textit{Invented}, \textit{Subjective}, \textit{Unverifiable}. This classification aids in the precise identification and correction of hallucinations in LLM outputs, improving model accuracy and text quality.
However, their classification is subjective and lacks standardized criteria, especially for \textit{Invented}, \textit{Subjective}, and \textit{Unverifiable} types, and lacks linguistic support. This section redefines judgement criteria of these hallucination types to clarify their concepts

We first coarsely divide the hallucinations of LLMs into two types:

(1) \textbf{\textit{Verifiable Error}}: Statements that are directly contradicted by factual evidence, meaning the ground truth can be determined.
 
(2) \textbf{\textit{Unverifiable Info}}: Statements that cannot be directly supported or refuted by factual evidence.

Secondly, hallucinations can be classified based on their editability: those that can be corrected through editing tasks and those that cannot:

(1) \textbf{\textit{Modifiable Error}} Verifiable statements that can usually be corrected by modifying a small part of the sentence. These are typically errors at the phrase level or below.

(2) \textbf{\textit{Non-modifiable Error}} Statements that are either unverifiable or, if verifiable, contradict factual evidence. Such errors occur in the premise of the sentence rather than in the sentence itself. Therefore, they cannot be corrected by simply modifying a small part of the sentence and are typically considered errors at the sentence level.

Base on these classification criteria, we redefine the 6 fine-grained hallucination types as Table \ref{halluc_types}.

\begin{table*}
  \centering
  \caption{Fine-grained hallucination types}
  \label{halluc_types}
  \resizebox{\linewidth}{!}{
  \begin{tabular}{c|p{0.9\linewidth}}
      \hline
      Hallucination     & Judgment Criteria \\ 
      \hline 
      \multirow{2}{*}{\textbf{\textit{Entity}}} &  Verifiable and Modifiable, and the contradiction is caused by entities.   \\ & Example:  Italo Calvino was a \sout{football player} novelist.   \\
      \hline 
      \multirow{2}{*}{\textbf{\textit{Relation}}} &  Verifiable and Modifiable, and the contradict is caused by a semantic relation error between objects, which may concern the verbs, tenses, or pronouns.  \\ & Example:  The cat was \sout{barking} meowing loudly at the passing cars.   \\
      \hline 
      \multirow{2}{*}{\textbf{\textit{Contradictory}}} &  Verifiable but Non-modifiable. The entire statement is contradicted by factual evidence, indicating that the errors arise from the false foundational premise of the sentence, not just individual words or phrases. Correcting such errors likely requires a complete reconstruction of the sentence, adjusting both content and structure to align with the truth. \\ & Example:  Calvino is widely considered the GOAT in basketball.   \\
      \hline 
      \multirow{2}{*}{\textbf{\textit{Invented}}} &  Unverifiable and Non-modifiable. The sentence refers to concepts that are clearly impossible in the real world, such as those that violate historical facts or the laws of physics. It may also contain purely fictional entities that do not exist in reality or in any known fictional works like novels, movies, or games. \\ & Example:  Calvino wrote a novel named \textit{Iron Hammer and Water}   \\
      \hline 
      \multirow{2}{*}{\textbf{\textit{Subjective}}} &  Unverifiable and Non-modifiable. Subjective sentences often include adjectives conveying strong personal emotions (e.g., 'terrible,' 'heartbreaking'), modal adverbs, superlatives, or expressions of totality, all of which are unverifiable and lack truth value~\cite{ochs1989language}. Subjectivity should come from the LLM itself rather than from others. \\ & Example:  Calvino is the best writer in the world.   \\
      \hline 
      \multirow{2}{*}{\textbf{\textit{Unverifiable}}} &  Unverifiable and Non-modifiable. The statements are not widely recognized or lack substantial evidence. They often involve speculation based on preliminary or incomplete data and personal privacy issues. While these statements may be logically consistent, they do not contradict known facts or the actual context. \\ & Example: Calvino liked to have a small pudding after dinner. \\
      \hline
  \end{tabular}
  }
\end{table*}

\section{Progressive Fine-grained Model Editor (PFME)}

\begin{figure*}[!t]
  \centering
  \includegraphics[width=1.0\textwidth]{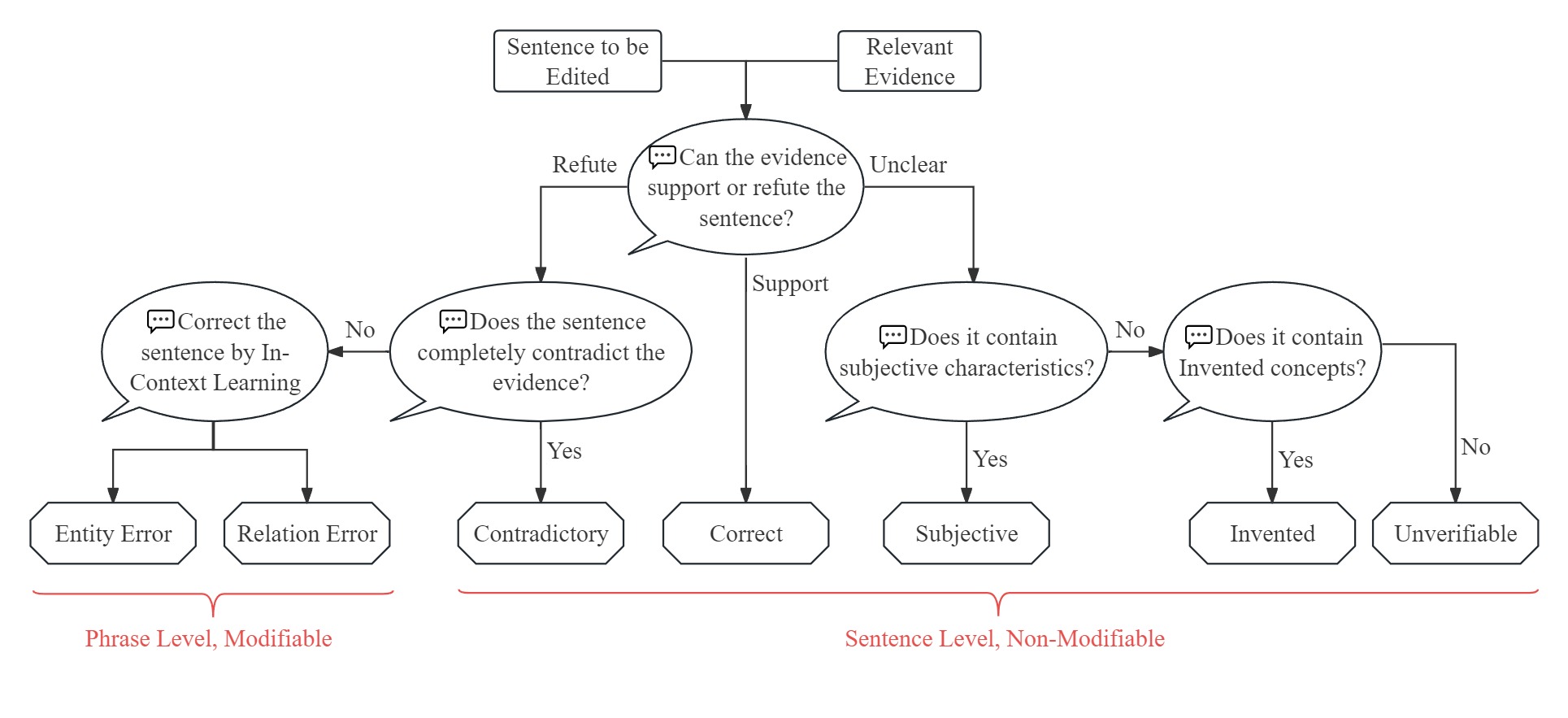}%
  \hfil
  \caption{Progressive Fine-grained Model Editor (PFME)'s Detection and Editing Module}
  \label{fig_edit_module}
  \end{figure*}

PFME framework comprises two modules: the \textit{retrieval} module and the \textit{detection and editing} module.
For the document to be edited: $D=\{s_1, s_2, ..., s_n\}$, the \textit{retrieval} module will retrieve text chunks $EV = \{ev_1, ev_2, ..., ev_k\}$ as evidence, and then the \textit{detection and editing} module will determine the specific hallucination type $err\_type_i$ of each sentence $s_i$ through a decision tree.

The PFME's innovative approach simplifies complex multi-task problems by breaking them down into more manageable, independent sub-tasks. Each sub-task is designed to detect and correct specific types of hallucinations. This strategy reduces reliance on the model's context learning capabilities and allows for the development of specialized detection and editing techniques tailored to various hallucination types. Furthermore, the PFME framework features a modular design and an ICL strategy, enhancing its adaptability and scalability for evolving tasks.

\subsection{Real-time Fact Retrieval module}

The \textit{retrieval} module of PFME consists of two phases: Recall and Ranking.

The Recall phase includes the following steps:
(1) Use a LLM for named entity recognition (NER) task on the text scheduled for editing, identifying key entities. To minimize ambiguity, it's crucial to append a brief definition after identifying each key entity.
(2) Utilize the identified key entities to retrieve relevant titles from the MediaWiki search engine, accessing the core content of relevant Wikipedia articles. Additionally, extract infobox data associated with the entities to enhance the informational breadth of the retrieved evidence text. During this phase, structured infobox data is converted into declarative statements to streamline subsequent processing.

The Ranking phase involves the following steps:
(1) Segment the retrieved evidence text to align with the editing model's context window constraints, ensuring that the semantic coherence of the text is preserved and that critical information is not truncated.
(2) Convert the segmented evidence text and the designated editing sentence into an embedded format, projecting them into a vector space.
(3) Use the retrieval model to determine the similarity between the sentence being reviewed and the evidence text segments. Then, apply the k-nearest neighbor (k-NN)~\cite{guo2003knn} algorithm to extract the k most relevant evidence text segments related to the sentence, which will serve as input for subsequent editing tasks.

\subsection{Hallucinations Detection and Edit Module}

In this module, PFME first employs InstructGPT (gpt-3.5-turbo-instruct) to break down the passage into individual sentences (including clauses), establishing a basis for detailed processing. It then proceeds with a two-stage retrieval process for each sentence, initially identifying the top $k$ relevant evidence chunks and integrating edited context from the text repository as trustworthy factual evidence.

Following this, PFME enters a decision-making phase to assess the accuracy of each sentence. If evidence suggests a discrepancy with the facts, PFME identifies the sentence as containing a verifiable error. It determines whether the entire sentence or parts of it conflict with the evidence. If the entire sentence is incorrect, it's flagged as a \textit{Contradictory} statement. If only parts are wrong, the sentence is directed to the editable branch.

Within editable branch, PFME utilizes factual evidence to modify and rectify sentences to ensure their accuracy. It highlights the erroneous parts of the original sentence, proposes corrective content, and identifies error types. If the error stems from entity inaccuracies, an \textit{Entity} tag is appended to that portion; if it arises from semantic relationship errors such as verbs, pronouns, tense, etc., a \textit{Relation} tag is added to that segment. Multiple \textit{Entity} and \textit{Relation} tags can coexist within the same sentence.

For sentences lacking evidence-based support, PFME categorizes them as unverifiable. It further distinguishes \textit{Subjective} statements, \textit{Invented} concepts, and other \textit{Unverifiable} claims. \textit{Subjective} statements are identified using subjective adjectives like "terrible" or "best". \textit{Invented} concepts refer to entities or ideas that don't exist or concepts contradicting common sense and knowledge without evidence. Statements aligning with established knowledge but lacking direct evidence to categorize them as \textit{Unverifiable}.

For sentences verifiably correct, PFME categorizes them as such and returns them without edits.

Ultimately, PFME applies appropriate editing strategies to all categorized sentences, such as conducting local edits for \textit{Entity} and \textit{Relation}, directly removing sentences flagged as \textit{Contradictory} and \textit{Invented}, highlighting warnings for \textit{Subjective} and \textit{Unverifiable}; and storing revised sentences in the text repository for subsequent assessment.

\section{Experiment}

\paragraph{Models}

For \textit{Edit Model}, we leverage two prominent edit models, namely ChatGPT (version gpt-3.5-turbo-0125)~\cite{openai2024chatgpt} and Llama3 (Meta-Llama3-8B-Instruct)~\cite{llama3modelcard}. ChatGPT is renowned for its conversational abilities and text generation, while Llama3 is selected for its instructive capabilities, which are beneficial for fine-grained editing tasks.
For \textit{Retrieval Model}, we use gte-large-en-v1.5~\cite{li2023towards} for the retrieval of evidence, leveraging its capability to process long-form text and extract relevant contextual information from vast datasets.

\paragraph{Benchmark and metrics}

For \textit{Detection Task}, we utilize a subset of 500 examples from the FAVA training dataset. The rationale behind this selection is the absence of a readily available Benchmark file from FAVA. The performance is gauged using the following metrics:
(1) \textit{Individual hallucination F1-score}: Assesses the model's ability to correctly identify and classify individual hallucination types.
(2) \textit{Overall Accuracy (OA) F1-score}: Reflects the model's overall performance across all hallucination types, considering precision and recall weighted by category proportions.
(3) \textit{Binary Prediction (Bi) F1-score}: Measures the model's proficiency in distinguishing between fact and hallucination.

For \textit{Edit Task}: we use the original biography generation task proposed with FActScore~\cite{min2023factscore}. These datasets serve to verify the efficacy of PFME in enhancing the factuality of text post-detection and editing processes.The primary metric for this phase is \textit{FActScore}: A metric designed to evaluate the factual accuracy of generated text. This model-based metric prompts ChatGPT and InstructGPT (gpt-3.5-turbo-Instruct) to decompose a response into a set of atomic facts and verify factuality for each using passages from a designated Wikipedia article.
Since FActScore evaluates the factuality of the text, both \textit{Unverifiable} and \textit{Subjective} are deemed non-factual in the FActScore assessment process. Therefore, for the FActScore task, texts labeled as \textit{Unverifiable} and \textit{Subjective} will be deleted along with those labeled as \textit{Contradictory} and \textit{Invented}.

\paragraph{Baseline}
Given the novelty of this field and the lack of relevant baselines, we mainly selected the most pertinent research~\cite{mishra2024fine-fava} as a baseline to compare with our method. Furthermore, our approach is not only applicable to large language models (LLMs) but also provides significant assistance in proofreading human-generated text.

For simplicity, we denote the editor parameters as "Editing Method @ Editing Model @ Evidence Count". For both detection and editing tasks, we evaluate the performance of various baseline models. Specifically, we utilize the \textbf{FavaP@ChatGPT@n} editor, which is based on the ChatGPT model~\cite{openai2024chatgpt}. This model is enhanced with Fava's prompt~\cite{mishra2024fine-fava} and augmented with n retrieved evidence documents to supplement the prompt. Additionally, we include Llama3, also augmented with n retrieved evidence chunks. Notably, n=0 indicates the absence of retrieved evidence. Following Fava's setup, the chunk size is set to 600.

Our experiment uses two editing methods: FavaP and PFME. For PFME, we use Llama3 as the editing model; for FavaP, we compare ChatGPT and Llama3 as editing models. The specific evidence counts are set to 0 and 5, with 0 indicating no external evidence is introduced. Additionally, since our experiments found that FavaP performed best with an OA metric when the evidence count was 1, we included additional results with an evidence count of 1 to ensure fairness.

\subsection{Main Results}

\begin{table*}
\small
  \centering
  \caption{Overall fine-grained detection performance. (F1-score)}
  \label{4_detect_result_f1}
  \resizebox{\linewidth}{!}{
  \begin{tabular}{c|cccccccc}
      \hline
      Editor               & Entity & Relation & Contradictory & Invented & Subjective & Unverifiable &  OA  &   Bi   \\ 
      \hline 
      FavaP@ChatGPT@0       &  28.4  &  8.0   &  \textbf{26.4}  &  9.1   &  30.2  &  14.1  &  22.1  &   56.7    \\ 
      FavaP@ChatGPT@1       &  29.8  &  9.2   &  26.0  &  11.3  &  26.3  &  13.2  &  22.2  &   53.5    \\ 
      FavaP@ChatGPT@5       &  30.6  &  7.4   &  18.1  &  9.1   &  20.6  &  11.6  &  19.2  &   51.6    \\ 
      \hline 
      FavaP@Llama3@0        &  24.9  &  11.4  &  15.4  &  15.6  &  8.7   &  13.0  &  18.5  &   50.9    \\ 
      FavaP@Llama3@1        &  28.0  &  15.1  &  15.2  &  11.4  &  7.8   &  15.8  &  20.9  &   51.7    \\ 
      FavaP@Llama3@5        &  28.9  &  9.8   &  9.8   &  1.4   &  9.2   &  5.3   &  17.2  &   48.1    \\ 
      \hline 
      PFME@Llama3@0         &  11.7  &  11.1  &  7.4   &  22.2  &  24.8  &  6.1   &  15.6  &   54.0    \\ 
      PFME@Llama3@1         &  32.5  &  22.5  &  19.4  &  34.6  &  31.4  &  19.2  &  28.1  &   66.1    \\ 
      PFME@Llama3@5         &  \textbf{34.9}  &  \textbf{29.8}  &  18.7  &  \textbf{34.9}  &  \textbf{33.6}  &  \textbf{24.2}  &  \textbf{30.9}  &   \textbf{69.6}    \\ 
      \hline
  \end{tabular}
  }
\end{table*}

\paragraph{Detection Task}
As shown in Table \ref{4_detect_result_f1},
the PFME model achieved the best overall performance metrics OA (Overall Accuracy) and Bi (Bivariate classification for factual errors) across all settings, significantly outperforming existing methods.

Comparison  shown in Table \ref{4_detect_result_compare} indicates:
(1) \textit{Different Methods}: PFME@Llama3 outperformed FavaP@Llama3 with OA and Bi improvements of 10.0 and 17.9 percentage points (pp), respectively, corresponding to enhancements of 47.8\% and 34.6\%.
(2) \textit{Different Methods and Different Editing Models}: PFME@Llama3 achieved OA and Bi improvements of 8.7 pp and 12.9 pp over FavaP@ChatGPT, corresponding to enhancements of 39.2\% and 22.8\%.
(3) \textit{With/Without evidence setting}: The OA and Bi scores for FavaP@ChatGPT and FavaP@Llama3 showed little difference before and after introducing evidence, with potential performance declines. However, PFME@Llama3@5 showed OA and Bi improvements of 15.3 pp and 15.6 pp over PFME@Llama3@0, corresponding to enhancements of 98.1\% and 28.9\%.

\begin{table*}
\small
  \centering
  \caption{Overall fine-grained detection performance comparison. (F1-score)}
  \label{4_detect_result_compare}
  \begin{tabular}{c|cc}
      \hline
      Editor       & OA Improvement    & Percentage Increase (OA)  \\ 
      \hline 
      PFME@Llama3 vs. FavaP@Llama3     &  10.0 pp &  47.8\%   \\ 
      PFME@Llama3 vs. FavaP@ChatGPT    &  8.7 pp  &  39.2\%   \\ 
      PFME@Llama3@5 vs. PFME@Llama3@0  &  15.3 pp &  98.1\%   \\   
      \hline
      \hline
      Editor       & Bi Improvement & Percentage Increase (Bi) \\ 
      \hline 
      PFME@Llama3 vs. FavaP@Llama3   &  17.9 pp  &  34.6\%   \\ 
      PFME@Llama3 vs. FavaP@ChatGPT  &  12.9 pp  &  22.8\%    \\ 
      PFME@Llama3@5 vs. PFME@Llama3@0  & 15.6 pp	 &   28.9\%   \\ 
      \hline 
  \end{tabular}
\end{table*}

Furthermore, we delve into the implications of these findings:
(1) \textit{Model and method comparison}: The PFME method based on Llama3 significantly outperformed FavaP@ChatGPT and FavaP@Llama3 in both OA and Bi metrics. This indicates the superior capability of the PFME method in handling evidence and error detection.
(2) \textit{Impact of the amount of evidence}: Without evidence, PFME's performance was subpar as it is designed specifically for scenarios with factual evidence. However, even with the introduction of a single piece of evidence, PFME's performance far exceeded that of FavaP. This demonstrates PFME's high dependency on evidence, with significant performance boosts when evidence is present.
(3) \textit{Influence of the editing model}: FavaP@Llama3 performed worse than FavaP@ChatGPT, possibly due to Llama3's slightly inferior performance in specific editing tasks compared to ChatGPT. However, when combined with the PFME method, Llama3 exhibited significant advantages, likely due to the PFME method's ability to better leverage Llama3's model structure and capabilities.

\paragraph{Edit Task}
The experimental results in Table \ref{4_edit_result_factscore} show the PFME editing method achieved the highest performance scores with 5 evidence counts. Applying PFME to the original text (No Edit) increased the FActScore by 3.7 pp, marking a 4.9\% improvement. PFME also showed notable enhancement in factuality over No Edit with 1 evidence count. In contrast, the baseline method did not demonstrate significant improvement across different evidence counts compared to No Edit and, in some instances, led to a decline in FActScore.

\begin{table}
  \centering
  \caption{Overall edit performance. (FActScore)}
  \label{4_edit_result_factscore}
  \begin{tabular}{c|c}
      \hline
      Editor       & FActScore \\ 
      \hline
      No Edit      &  76.0    \\ 
      FavaP@0      &  74.2     \\ 
      FavaP@1      &  76.2     \\ 
      FavaP@5      &  74.9     \\ 
      \hline
      PFME@1       &  78.6  \\ 
      PFME@5       &  \textbf{79.7}     \\ 
      \hline
  \end{tabular}
\end{table}

In addition, we have recorded the processing time of the detection task in Table \ref{4_detect_time}. It is noteworthy that the quantity of evidence doesn't impact operational speed in the code implementation. The results illustrate that even though the parameter scale of PFME@Llama3 is smaller and computational power is limited, it exhibits the best inference performance and highest operational efficiency across all settings. Furthermore:

(1) For the FavaP setting, Llama3 is slower than ChatGPT because despite Llama3 having a smaller parameter scale, it operates only on a single A100 40GB GPU, thus its computational power is limited. Therefore, under the same settings, the single A100 40GB GPU running Llama3 is expected to be slower than ChatGPT in terms of invocation speed.
(2) Similarly, using Llama3 as the editing model for both FavaP and PFME, the processing speed of FavaP@Llama3 is inferior to PFME@Llama3. This is because in the PFME processing pipeline, neither sentence-level hallucinations nor correct sentences need token outputs from the editing model; they are directly labeled based on classification results, thereby reducing the workload of the editing model output.

Combining the above points, along with the operational efficiency of PFME@Llama3 and FavaP@ChatGPT, it can be observed that the PFME framework enables PFME@Llama3 to surpass the operational speed of ChatGPT with lower computational power.

Moreover, in comparing the operational speeds of PFME@Llama3 and PFME@ChatGPT, it is evident that PFME tends toward a high-frequency, low-throughput request pattern. Therefore, using ChatGPT as the editing model is primarily constrained by the network interface access rate, while FavaP tends toward a low-frequency, high-throughput request pattern, mainly constrained by computational power and contextual limitations.

\begin{table*}
  \centering
  \caption{Detection Task Process Time}
  \label{4_detect_time}
  \begin{tabular}{c|cccccccc}
      \hline
      Editor         & Time Cost(min) & Process Speed(sec/sample)  & OA(f1-score)  & Bi(f1-score)  \\ 
      \hline 
      FavaP@ChatGPT  &  28.53    &      3.42            & 22.2  & 53.5 \\ 
      FavaP@Llama3   &  34.20    &      4.10            & 20.9  & 51.7 \\ 
      \hline 
      PFME@ChatGPT   &  143.75   &      17.25           & 15.5  & 58.8 \\ 
      PFME@Llama3    &  \textbf{27.62}  &  \textbf{3.31}   & \textbf{30.0}  & \textbf{67.3} \\ 
      \hline
  \end{tabular}
\end{table*}

\section{Analysis}  
To thoroughly explore how the quantity of evidence influences PFME performance, we conduct a series of ablation experiments. These experiments are designed to assess the impact of varying amounts of external knowledge evidence on PFME's effectiveness in fine-grained hallucination classification and editing tasks. 

\subsection{Detection Task Ablation: Evidence Num} 
In the fine-grained hallucination detection task, ablation experiments show that incorporating external knowledge evidence positively affects PFME performance. As the number of evidence chunks increases from 1 to 5 (PFME@1 to PFME@5), overall performance metrics OA and Bi enhance, as shown in Table \ref{4_detect_ablation_evi_num}. Visual analysis in Figure \ref{fig_ablation_similarity} (the "ret" line) indicates that OA and Bi metrics initially rise but then decline as the evidence quantity increases from PFME@1 to PFME@10. The PFME@5 setup achieves the best F1 score.

Notably, unverifiable types of hallucinations, such as \textit{Invented}, \textit{Subjective}, and \textit{Unverifiable}, exhibit trends similar to overall metrics (OA and Bi). This similarity arises because improved identification of verifiable hallucinations reduces the likelihood of misclassifying them as unverifiable types, thereby indirectly enhancing the accuracy of classifying unverifiable hallucinations. In summary, the results emphasize the significance of the quantity of external knowledge evidence for enhancing PFME performance and highlight the importance of factual evidence in accurately classifying unverifiable types within the PFME framework.

\subsection{Edit Task Ablation: Dataset} 
In the FActScore dataset, the ChatGPT dataset initially received a high score before any edits, as shown in Table \ref{4_edit_ablation_dataset}. As a result, PFME's improvements on the ChatGPT dataset were minimal, with the highest score increasing by just 4.6 pp and factuality improving by 6.1\% compared to No Edit.
To evaluate PFME's reliability and generalization ability in editing tasks, we analyze the Alpaca 13B dataset from FActScore. Like the ChatGPT dataset, the Alpaca 13B dataset consists of biographies generated by the Alpaca 13B for 500 individuals.

In our ablation study, we test PFME@1 to PFME@10 to assess the impact of using different numbers of evidence chunks on editing the Alpaca 13B dataset. We also compare its performance to PFME's performance on the ChatGPT dataset using the same number of evidence chunks. The detailed results are presented in Table \ref{4_edit_ablation_dataset}, showing that PFME@4 achieves the highest FActScore on the Alpaca 13B dataset, with the score increasing by 16.2 pp, and improving factuality by 32.7\% from the original 49.5.
Specifically, PFME performs best using 4 evidence chunks on the Alpaca 13B dataset and 7 chunks of evidence on the ChatGPT dataset. This suggests an optimal balance between evidence richness and redundancy. Since this balance varies across different datasets, future research should explore mechanisms for selecting evidence based on quality, diversity, and relevance to determine the optimal number of evidence chunks universally.

\begin{table}
  \centering
  \caption{Edit Task: Ablation-Dataset (FActScore)}
  \label{4_edit_ablation_dataset}
  \begin{tabular}{c|c|c}
      \hline
      Editor       & ChatGPT & Alpaca 13B   \\ 
      \hline
      No Edit     &  76.0  &   49.5        \\ 
      PFME@1      &  78.6  &   64.5     \\
      PFME@2      &  78.3  &   64.6     \\
      PFME@3      &  80.1  &   65.0     \\
      PFME@4      &  80.1  &   \textbf{65.7}     \\
      PFME@5      &  79.7  &   65.1     \\
      PFME@6      &  80.1  &   64.9     \\
      PFME@7      &  \textbf{80.6}  &   64.6     \\
      PFME@8      &  80.3  &   61.7     \\
      PFME@9      &  79.6  &   56.7     \\
      PFME@10     &  80.0  &   55.6     \\
      \hline
  \end{tabular}
\end{table}

\subsection{More Ablation Experiments}
We conduct additional ablation experiments as detailed in the Appendix section. 

In Appendix \ref{appendix: Similarity Ranking Method}, we perform ablation experiments to evaluate how various retrieval methods affect hallucination detection and editing. We find that ranking candidate evidence with retrieval model embedding similarity provides a more consistent and superior performance, while an optimal balance point for evidence quantity may vary across datasets. Meanwhile, excessive irrelevant evidence can impair performance by causing information overload and exceeding context limits, highlighting the necessity for effective evidence selection.

In Appendix \ref{appendix: Retrieval Level}, we design ablation experiments to examine the influence of retrieval levels, comparing document granularity with sentence granularity. We find that document-level retrieval outperforms sentence-level retrieval in most categories, providing more accurate evidence selection, while being more efficient in practical applications, especially with large document volumes. However, in the Contradictory classification, sentence-level retrieval excels, suggesting its advantage in handling specific hallucination types.

\section{Conclusion}

We introduce the Progressive Fine-grained Model Editor (PFME), a framework designed to detect and correct fine-grained hallucinations in large language models. PFME decomposes complex tasks into manageable sub-tasks and uses specialized prompts for various hallucination types, enhancing adaptability, scalability, and readability. Our experiments show PFME outperforms existing methods in detection and editing, particularly in Overall Accuracy and Binary Prediction Accuracy, even with limited computational resources. Despite its effectiveness, there is room for improvement in hallucination classification and reasoning capabilities. Future work will explore advanced prompt engineering methods and await the final version of the FavaBench test benchmark.

\section{Limitations}

The PFME framework has yielded positive results but has also shown some limitations. Firstly, the editing module uses hard prompts, which may limit its performance improvement potential. In the future, we plan to optimize it using advanced techniques such as fine-tuning small models or P-tuning. Lastly, the current classification method is still in its early stages and requires further research. We aim to incorporate more linguistic knowledge to enhance the comprehensiveness and clarity of the classification.

\section{Ethics Statements}

Our research focuses on utilizing a fine-grained hallucination taxonomy to identify and correct hallucinations in text. However, experiments have revealed that the PFME framework may still miss or mislabel hallucinations generated by LLMs.

We assessed our model's detection capabilities using a ChatGPT-generated, privacy-compliant training set by Mishra~\cite{mishra2024fine-fava}. For editing performance, we utilized the open-source FActScore tool~\cite{min2023factscore}, which relies on Wikipedia-based, non-intrusive datasets.

AI Assistant Statement: As the authors are not native English speakers, we have utilized  ChatGPT to check grammar and spelling errors and to refine the original expressions, which is purely in assistance with the language of the paper.

\bibliography{latex/PFME}

\appendix

\section{Detection Task Ablation: Similarity Ranking Method} 
\label{appendix: Similarity Ranking Method}

The PFME method optimizes the editing process in evidence retrieval through two core steps. Firstly, the method utilizes a retrieval model to calculate the cosine similarity between the sentence to be edited and all evidence texts, selecting the top 10 most relevant segments as candidate evidence. Secondly, PFME further refines this selection by ranking the candidate evidence to better match the sentence to be edited.

We conduct ablation experiments to assess the impact of different retrieval methods on hallucination detection and editing. We compare four similarity ranking methods:

(1) \textbf{Retrieval Similarity (ret)}: Cosine similarity is calculated using embeddings output by the retrieval model.
(2) \textbf{SpaCy Similarity (nlp)}: Integrates entity matching, extracting entities and calculating their cosine similarity using SpaCy.
(3) \textbf{Fusion Similarity (fus)}: Combines the above methods using the formula fus = (ret + nlp) / 2.
(4) \textbf{Random Selection (rnd)}: No similarity calculation is performed, randomly selects evidence from candidate evidence.

\begin{figure}[!t]
  \centering
  \includegraphics[width=3.0in]{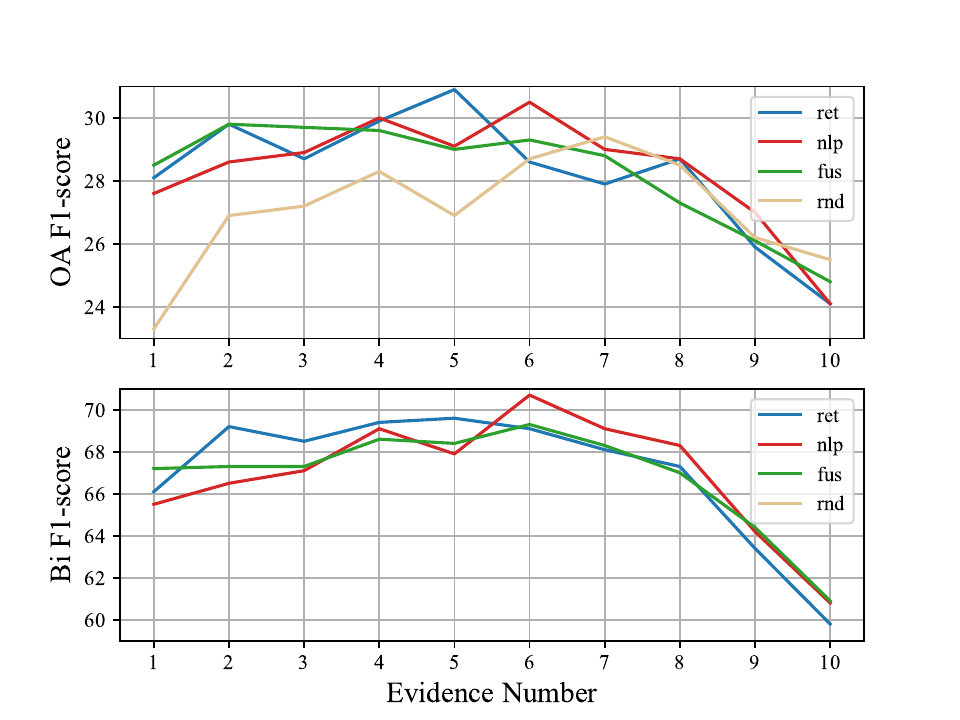}
  \hfil
  \caption{Ablation: Similarity}
  \label{fig_ablation_similarity}
\end{figure}

In this experiment, we tested the evidence count range from 1 to 10, named "PFME@evidence\_count@ranking\_method". Due to space limitations, we show the experimental results for evidence counts of 2, 4, 5, 6, and 9 in Table \ref{4_detect_ablation_evi_sim}, while the complete experimental results are presented in line graph form in Figure \ref{fig_ablation_similarity}. The following is a detailed analysis of the experimental results:

\subsection{Evidence Quantity and Performance Balance}
(1) \textbf{Singular Performance}: At an evidence count of 5, the ret method performs best in terms of the OA metric across all settings; however, when the evidence count increases to 6, the nlp method achieves the highest Bi metric performance across all settings.
(2) \textbf{Comprehensive Performance}: Within the evidence count range of 2 to 5, the ret method exhibits the most excellent performance in the overall accuracy (OA) and binary prediction accuracy (Bi) composite evaluations. At evidence counts of 1 or 10, the fus method performs best in the composite evaluation of OA and Bi; otherwise, the nlp method demonstrates the best overall performance.

In summary, the ret ranking method can more stably achieve better performance for further sorting of candidate evidence. Additionally, there may be a point in the number of evidence where there's enough information for editing without causing redundancy or conflicts. Since this balance point differs between the two datasets, future research could analyze evidence selection mechanisms, evidence quality, diversity, and relevance to determine if there exists a universally optimal balance point for evidence quantity.

\subsection{Necessity of Evidence Similarity Ranking Methods}

(1) \textbf{Impact of Low Evidence Quantity}: With a low evidence count, the performance of randomly selected candidate evidence is notably lower compared to results utilizing similarity sorting methods.
(2) \textbf{Performance Convergence at High Evidence Quantity}: With a high evidence count, the model's performance tends to converge, primarily due to the task setup: considering a total of 10 candidate evidence texts, when most of them are taken into account, the differences between different similarity calculation methods mainly manifest in the sorting of evidence texts.
(3) \textbf{Performance Decline Due to Excessive Evidence Quantity}: Beyond 9 evidence counts, the F1-score of verifiable hallucination types sharply drops as it exceeds the context window length of Llama3 in some subtasks (such as \textit{Entity} and \textit{Relation}). Meanwhile, at 5 to 8 evidence counts, OA performance fluctuates with a slow decline, while Bi performance remains stable, indicating information overload diluting effective information within the context.

In summary, introducing excessive irrelevant evidence may lead to information overload, dilution of effective information within the context, or exceeding the context window limit, thereby affecting overall performance. Therefore, within a limited context window, effective methods for selecting relevant evidence texts are needed to select the most relevant evidence while avoiding irrelevant evidence. Effective sorting methods can significantly improve model task performance with fewer evidence quantities.

\begin{table*}
  \centering
  \caption{Detection Task: Ablation-Similarity (Metric: F1-score)}
  \label{4_detect_ablation_evi_sim}
  \resizebox{\linewidth}{!}{
  \begin{tabular}{c|cccccccc}
      \hline
      Editor        & Entity & Relation & Contradictory & Invented & Subjective & Unverifiable &   OA  &   Bi   \\ 
      \hline 
      PFME@2@ret    & 31.3     & 27.0     &     23.1       &   35.2    &      30.9    &     23.5     & 29.8 & 69.2 \\
      PFME@2@nlp    & 29.1     & 25.0     &     22.4       &   34.6    &      33.8    &     19.5     & 28.6 & 66.5 \\
      PFME@2@fus    & 31.1     & 25.8     &     22.9       &   37.7    &      32.8    &     22.0     & 29.8 & 67.3 \\
      PFME@2@rnd    & 24.9     & 24.8     &     19.6       &   32.9    &      30.4    &     20.8     & 26.9 & 63.7 \\
      \hline                                       
      PFME@4@ret    & 31.6     & 27.2     &     19.9       &   \textbf{37.0}    &      33.0    &     25.8     & 29.9 & 69.4 \\
      PFME@4@nlp    & 31.0     & 25.0     &     \textbf{25.4}       &   35.4    &      31.2    &     \textbf{26.5}     & 30.0 & 69.1 \\
      PFME@4@fus    & 31.8     & 25.0     &     21.2       &   35.6    &      32.1    &     26.4     & 29.6 & 68.6 \\
      PFME@4@rnd    & 26.2     & 23.2     &     23.6       &   33.0    &      33.9    &     25.1     & 28.3 & 65.5 \\
      \hline                                       
      PFME@5@ret    & \textbf{34.9}     & \textbf{29.8}     &     18.7       &   34.9    &      33.6    &     24.2     & \textbf{30.9} & 69.6 \\
      PFME@5@nlp    & 29.8     & 24.2     &     20.3       &   34.6    &      36.1    &     21.4     & 29.1 & 67.9 \\
      PFME@5@fus    & 30.7     & 25.4     &     20.2       &   33.7    &      35.2    &     21.2     & 29.0 & 68.4 \\
      PFME@5@rnd    & 27.2     & 22.2     &     18.6       &   30.7    &      30.4    &     23.3     & 26.9 & 66.3 \\
      \hline                                       
      PFME@6@ret    & 31.4     & 24.8     &     19.7       &   31.5    &      32.8    &     22.2     & 28.6 & 69.1 \\
      PFME@6@nlp    & 30.0     & 27.4     &     18.8       &   36.9    &      \textbf{36.3}    &     23.0     & 30.5 & \textbf{70.7} \\
      PFME@6@fus    & 31.5     & 25.3     &     18.6       &   33.4    &      35.0    &     22.4     & 29.3 & 69.3 \\
      PFME@6@rnd    & 27.8     & 24.6     &     17.1       &   33.2    &      35.4    &     22.3     & 28.7 & 68.0 \\
      \hline                                       
      PFME@9@ret    & 25.5     & 20.9     &     19.8       &   30.4    &      31.4    &     22.9     & 25.9 & 63.4 \\
      PFME@9@nlp    & 24.2     & 23.8     &     19.2       &   34.0    &      30.2    &     25.4     & 27.0 & 64.2 \\
      PFME@9@fus    & 24.4     & 23.3     &     18.1       &   32.3    &      30.1    &     23.1     & 26.1 & 64.4     \\
      PFME@9@rnd    & 23.1     & 22.5     &     18.5       &   29.1    &      32.2    &     27.2     & 26.2 & 62.9 \\
      \hline
  \end{tabular}
  }
\end{table*}

\section{Detection Task Ablation: Retrieval Level} 
\label{appendix: Retrieval Level}

Considering that the PFME method processes text at the sentence level granularity, we design ablation experiments to test the impact of retrieval levels based on document granularity versus sentence granularity on experimental results. The experimental setup is as follows:

(1) \textbf{Sentence-level Retrieval}: We employ a retrieval model to embed each sentence and individually compute its cosine similarity with all evidence text chunks. Subsequently, we select the top k text chunks based on cosine similarity as evidence. Given the content volume of the sentence to be edited, we set the text chunk size to 300. The experiment is named according to the convention "PFME@sent@evidence\_count," where the evidence count ranges from 1 to 10.
(2) \textbf{Document-level Retrieval}: This setup aligns with previous experiments. We utilize a retrieval model to embed the entire document to be edited and compute its cosine similarity with all evidence text chunks. When editing each sentence, we select the top k highest cosine similarity text chunks as evidence. Thus, the evidence cited for each sentence remains consistent. The text chunk size remains consistent with previous settings, set to 600. The experiment follows the naming convention "PFME@psg@evidence\_count," with the evidence count referencing the chunk information volume at the sentence level, set from 1 to 5.

The experimental results presented in Table \ref{4_detect_ablation_retrieval_level} indicate:
(1) \textbf{Performance Comparison}: According to Table \ref{4_detect_ablation_retrieval_level}, document-level retrieval demonstrates higher F1 scores than sentence-level retrieval in most classification categories, particularly in \textit{Relation}, \textit{Invented}, \textit{Subjective}, \textit{Subjective}, OA, and Bi metrics. This suggests that document-level retrieval generally provides more accurate evidence selection.
(2) \textbf{Efficiency}: At the point of maximal performance, the amount of evidence content required for document-level retrieval is comparable to that of sentence-level retrieval. However, the performance peak of document-level retrieval is higher, indicating that document-level retrieval may be more efficient in practical applications, especially when dealing with large volumes of documents.
(3) \textbf{Special Case}: In the \textit{Contradictory} classification, the performance of sentence-level retrieval significantly outperforms document-level retrieval. This may indicate that sentence-level independence might have an advantage when dealing with specific types of information.

\begin{table*}
  \centering
  \caption{Detection Task: Ablation-Retrieval Level (Metric: F1-score)}
  \label{4_detect_ablation_retrieval_level}
  \resizebox{\linewidth}{!}{
  \begin{tabular}{c|cccccccc}
      \hline
      Editor        & Entity & Relation & Contradictory & Invented & Subjective & Unverifiable &   OA  &   Bi   \\ 
      \hline 
      PFME@psg@1    &  32.5  &  22.5  &  19.4  &  34.6  &  31.4  &  19.2  &  28.1  &   66.1    \\ 
      PFME@psg@2    &  31.3  &  27.0  &  23.1  &  35.2  &  30.9  &  23.5  &  29.8  &   69.2    \\ 
      PFME@psg@3    &  32.3  &  25.7  &  21.9  &  34.4  &  32.5  &  18.0  &  28.7  &   68.5    \\ 
      PFME@psg@4    &  31.6  &  27.2  &  19.9  &  \textbf{37.0}  &  33.0  &  25.8  &  29.9  &   69.4    \\ 
      PFME@psg@5    &  34.9  &  \textbf{29.8}  &  18.7  &  34.9  &  \textbf{33.6}  &  24.2  &  \textbf{30.9}  &   \textbf{69.6}    \\ 
      \hline                                       
      PFME@sent@1   &  25.1  &  18.6  &  12.8  &  26.1  &  28.8  &  12.4  &  22.7  &   62.7    \\ 
      PFME@sent@2   &  28.3  &  21.8  &  15.2  &  29.1  &  28.7  &  13.1  &  24.4  &   64.3    \\ 
      PFME@sent@3   &  30.4  &  21.4  &  15.4  &  30.2  &  31.8  &  18.1  &  26.3  &   65.5    \\ 
      PFME@sent@4   &  31.0  &  23.6  &  18.7  &  33.6  &  31.9  &  19.0  &  27.9  &   66.3    \\ 
      PFME@sent@5   &  29.0  &  24.0  &  21.7  &  32.2  &  31.1  &  16.6  &  27.2  &   65.3    \\ 
      PFME@sent@6   &  33.4  &  23.0  &  23.0  &  31.1  &  29.7  &  22.7  &  28.7  &   66.0    \\ 
      PFME@sent@7   &  33.0  &  22.6  &  23.5  &  31.8  &  31.0  &  21.0  &  28.5  &   65.9    \\ 
      PFME@sent@8   &  32.3  &  27.0  &  \textbf{23.9}  &  31.0  &  30.6  &  17.8  &  28.5  &   66.6    \\ 
      PFME@sent@9   &  \textbf{35.8}  &  25.8  &  \textbf{23.9}  &  31.6  &  32.6  &  22.5  &  30.0  &   67.3    \\ 
      PFME@sent@10  &  33.3  &  27.8  &  21.9  &  31.6  &  30.8  &  21.9  &  29.3  &   67.0    \\ 
      \hline
  \end{tabular}
  }
\end{table*}

\section{Table: Detection Task: Ablation-Evidence Num (Metric: F1-score)} 

\begin{table*}
  \centering
  \caption{Detection Task: Ablation-Evidence Num (Metric: F1-score)}
  \label{4_detect_ablation_evi_num}
  \resizebox{\linewidth}{!}{
  \begin{tabular}{c|cccccccc}
      \hline
      Editor        & Entity & Relation & Contradictory & Invented & Subjective & Unverifiable &   OA  &   Bi   \\ 
      \hline 
      PFME@Llama3@1  &  32.5  &  22.5  &  19.4  &  34.6  &  31.4  &  19.2  &  28.1  &   66.1    \\ 
      PFME@Llama3@2  &  31.3  &  27.0  &  \textbf{23.1}  &  35.2  &  30.9  &  23.5  &  29.8  &   69.2    \\ 
      PFME@Llama3@3  &  32.3  &  25.7  &  21.9  &  34.4  &  32.5  &  18.0  &  28.7  &   68.5    \\ 
      PFME@Llama3@4  &  31.6  &  27.2  &  19.9  &  \textbf{37.0}  &  33.0  &  25.8  &  29.9  &   69.4    \\ 
      PFME@Llama3@5  &  \textbf{34.9}  &  \textbf{29.8}  &  18.7  &  34.9  &  \textbf{33.6}  &  24.2  &  \textbf{30.9}  &   \textbf{69.6}    \\ 
      PFME@Llama3@6  &  31.4  &  24.8  &  19.7  &  31.5  &  32.8  &  22.2  &  28.6  &   69.1    \\ 
      PFME@Llama3@7  &  30.3  &  25.0  &  17.8  &  32.7  &  33.0  &  22.6  &  27.9  &   68.1    \\ 
      PFME@Llama3@8  &  27.9  &  26.9  &  21.6  &  32.9  &  30.2  &  \textbf{27.0}  &  28.7  &   67.3    \\ 
      PFME@Llama3@9  &  25.5  &  20.9  &  19.8  &  30.4  &  31.4  &  22.9  &  25.9  &   63.4    \\ 
      PFME@Llama3@10 &  23.8  &  19.4  &  15.5  &  30.8  &  27.5  &  23.0  &  24.1  &   59.8   \\ 
      \hline
  \end{tabular}
  }
\end{table*}

\end{document}